\documentclass[final]{article} 
\usepackage{neurips_2024}
\usepackage{graphicx}

\usepackage{amsmath,amsfonts,bm}

\def\eqref#1{equation~\ref{#1}}

\def\1{\bm{1}}

\DeclareMathAlphabet{\mathsfit}{\encodingdefault}{\sfdefault}{m}{sl}
\SetMathAlphabet{\mathsfit}{bold}{\encodingdefault}{\sfdefault}{bx}{n}

\usepackage[utf8]{inputenc} 
\usepackage[T1]{fontenc}    
\usepackage{hyperref}       
\usepackage{url}            
\usepackage{booktabs}       
\usepackage{amsfonts}       
\usepackage{nicefrac}       
\usepackage{microtype}      
\usepackage{xcolor}         
\usepackage{multirow}
\usepackage{makecell}
\usepackage{arydshln}
\usepackage{wasysym}
\newcommand{\ours}{PORTAL}

\title{PORTAL: Scalable Tabular Foundation Models via Content-Specific Tokenization}

\author{Marco Spinaci \\
SAP France \\
\texttt{marco.spinaci@sap.com}
\And
Marek Polewczyk \\
SAP SE \\
\texttt{marek.polewczyk@sap.com}
\And
Johannes Hoffart \\
SAP SE \\
\texttt{johannes.hoffart01@sap.com}
\And
Markus C. Kohler \\
SAP SE \\
\texttt{markus.kohler01@sap.com}
\And
Sam Thelin \\
SAP SE \\
\texttt{sam.thelin@sap.com}
\And
Tassilo Klein \\
SAP SE \\
\texttt{tassilo.klein@sap.com}
}

\begin{document}

\maketitle

\begin{abstract}
  Self-supervised learning on tabular data seeks to apply advances from natural language and image domains to the diverse domain of tables. However, current techniques often struggle with integrating multi-domain data and require data cleaning or specific structural requirements, limiting the scalability of pre-training datasets.
  We introduce \ours{}\footnote{Code will be released at \url{https://github.com/SAP-samples/portal}} (Pretraining One-Row-at-a-Time for All tabLes), a framework that handles various data modalities without the need for cleaning or preprocessing. This simple yet powerful approach can be effectively pre-trained on online-collected datasets and fine-tuned to match state-of-the-art methods on complex classification and regression tasks. This work offers a practical advancement in self-supervised learning for large-scale tabular data.
\end{abstract}

\section{Introduction}

The introduction of BERT \cite{devlin-etal-2019-bert}, powered by the transformer architecture and enriched by extensive unstructured data from the internet, marked a transformative era in deep learning-based natural language processing. On the other hand, the impact of neural architectures on structured tabular data has been less significant. Tabular data, widely used in many enterprises, is a common target for machine learning applications but handling such data at scale presents challenges requiring scalable analytical solutions.

Currently, tree-based ensemble methods, particularly boosting techniques, excel in managing tabular data. Methods like XGBoost \cite{Chen:2016:XST:2939672.2939785} and CatBoost \cite{conf/nips/ProkhorenkovaGV18} generally outperform deep learning models in both effectiveness and resource efficiency, maintaining a strong dominance \cite{NEURIPS2022_0378c769}. 

Despite successes in natural language, neural network methods for tabular data often underperforms compared to gradient boosting methods, possibly due to the limited possibilities for transfer learning, stemming from stringent preprocessing requirements. While there are attempts to adapt transformer models for tabular data \cite{yang2024unitabe,ye2024crosstable,yan2024making}, few achieve results comparable to gradient boosting models, with successes generally confined to specific settings \cite{hollmann2023tabpfn,CARTE}.

To address these challenges, we propose \ours{} (\textbf Pretraining \textbf One-\textbf{R}ow-at-a-\textbf{T}ime for \textbf All tab\textbf{L}es), a transformer encoder-based model pre-trained on diverse tabular data sources, with data types including text, numbers, and dates. \ours{} demonstrates superior performance (especially on text-heavy datasets) with respect to both boosting (XGBoost) and transformer (CM2) approaches, while being comparable to, and sometimes better than, state-of-the-art approaches (CatBoost, AutoGluon, CARTE). Our core contributions are threefold:

\textbf{First,} we introduce an adaptable encoding structure that eliminates the need for normalization or special handling of missing values and outliers. \\
\textbf{Second,} we implement a pre-training scheme using masked cell modeling that generalizes effectively to fine-tuning, reducing the gap between pre-training and fine-tuning stages. \\
\textbf{Third,} we provide a comparative analysis with robust classical baselines to highlight performance gaps between traditional tree-based methods and our transformer-based approach.

\section{Related Work}\label{sec:related}

Several approaches have been proposed to pre-train models on large amounts of tabular data. Here, we highlight those most relevant to our work.

\cite{tapas} was among the first to adapt BERT's architecture to encode tables for question answering, yet it was constrained to smaller tables. \cite{XTab} developed a transformer model divided into an encoder, backbone transformer, and decoder, but with limited scalability as only the transformer was shared across pre-training datasets. Despite these advancements, traditional tree-based models like CatBoost still exhibited superior performance.

Following these, \cite{hollmann2023tabpfn} pre-trained a transformer on synthetic datasets targeting high accuracy and swift inference, though its utility was limited to smaller tables and specific classification tasks, making it less relevant to broader applications. \cite{yang2024unitabe} utilized TabUnit modules that treat table cells as key-value pairs, combining transformer and LSTM decoders, pre-training on a vast 13B-row dataset to slightly edge out XGBoost.

\cite{ye2024crosstable} advanced a positional-encoding-free transformer using masked cell modeling on large datasets, with nuances in text and numerical data handling that, while innovative, led only to marginal performance gains. Meanwhile, \cite{yak2023ingestables} introduced a "MapTransformer" architecture that converts table rows into feature embeddings sequences, albeit restricted to a select few large datasets for pretraining.

\cite{yan2024making} further explored transformer encoders pre-trained on discretized numerical features with the C4.5 algorithm but required labeled datasets, thus limiting its scalability. Finally, \cite{CARTE} presented a graph-based encoder forming data into star-shaped graphlets for fine-tuning, achieving state-of-the-art results with an ensemble of multiple models. Our experiments indicate that while ensemble methods are necessary for peak performance, satisfactory results are achievable with a standalone model setup.

\section{Method}\label{sec:method}
\noindent{\textbf{Preliminaries:}}

In this study, we implement an encoder-only transformer as our primary architecture, optimally adapted for generating embeddings from heterogeneous data types. The model takes feature values and dataset descriptors, such as column names, as inputs. Initially, dedicated type-specific encoders convert each type of data into fixed-length embeddings. These embeddings are then combined with column metadata information and transformed into a unified latent space, which feeds directly into the transformer's backbone. The output from the transformer is then decoded for each feature individually, allowing specific processing based on the feature type.

A key characteristic of our model is its capacity to handle data on a row-by-row basis without needing context from the rest of the table that the row comes from (or even to know if there was any other data beyond this one row). The training process of our model is executed in two main stages. In the first stage, the pre-training phase, the model sees randomly masked tokens, similarly to the process popularized by BERT pre-training \cite{devlin-etal-2019-bert}. In this step, the model sees a version of the data where content is masked, but column names are left unmasked, to ``prompt'' the model for completion. This is needed, since no other positional encoding is used, and therefore the model could otherwise not determine which masked value to predict for which cell when more than one cell is masked. The model is then fine-tuned on downstream tasks, typically consisting of either regression or classification. All of the weights are shared between the pre-training stage and the fine-tuning one except, possibly, the last output layer.

\begin{figure}[h]
\centering
\includegraphics[width=0.85\textwidth]{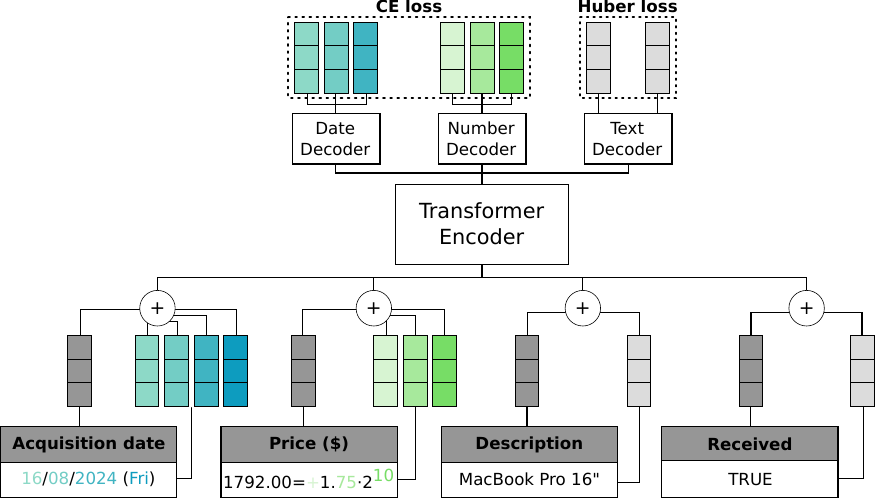}
\caption{\textbf{Schematic illustration of \ours{} architecture:} Example on a row with 4 columns. Blue cells denote custom encodings/decodings for date (day, month, year, day of the week, and holidays). Green cells correspond to numbers (sign, fraction, and exponent), and gray cells correspond to text embeddings. Dark gray cells are column name embeddings. All values are processed via a trainable linear/embedding layer before being aggregated by sum. In the output layer, similar decoding layers are applied before feeding the outputs to cross-entropy or Huber loss in training (see \ref{sec:decoding} for details).}
\label{fig:architecture}
\end{figure}

The conceptual framework of our method is structured into three sequential segments: encoding, backbone, and decoding. Detailed descriptions of each segment are elaborated in the subsequent sections of this paper. For a graphical representation of the proposed \ours{} approach, readers are directed to refer to Fig.~\ref{fig:architecture}, which visually outlines the structure and flow of our methodology.

Throughout the discussion, we denote the hidden dimension of the backbone transformer model by $d$.

\subsection{Encoding}
Depending on the type of input data, distinct encoding mechanisms are employed. Specifically, we introduce three specialized encodings tailored for the primary atomic data types: text, numerical values, and dates. For each cell in a table row, only one type of embedding—date, numerical, or string—is generated and summed with the header embedding derived from the column name or meta information.
At the end of the encoding process, each input row cell is represented by a single vector. Subsequently, all token embeddings are concatenated and passed to the backbone (in our case, transformer encoder).

\noindent{\textbf{Text:}}
For cells that include text data, we employ a pre-trained large language model (LLM) for sentence embedding. Similar techniques were also used in other papers, such as \cite{yak2023ingestables, CARTE, ye2024crosstable}; however, the details of how information from cell and content are mixed together differ. In \ours{}, we separately embed text content and column name via the same LLM. Both types of embeddings then undergo processing via two distinct trainable linear layers configured to deliver output in dimension $d$. This alignment is necessary to allow the model to distinguish between column name embeddings and text content embeddings.

\noindent{\textbf{Date:}}
For date cell values, the day, month, and year (clipped to the 1950-2050 range) are individually treated as integers and embedded in dimension $d$. As part of the automated feature engineering process, $d$-dimensional embeddings for the day of the week are incorporated, along with those of binary indicators that denote whether a date corresponds to a public holiday in any major countries and territories. These vectors are then summed up together with the embedding of the column name (followed by the same linear adaptation layer as above).

\noindent{\textbf{Numerical:}}
Numerical values in our model are expressed using scientific notation: $x = \pm \alpha \cdot 2^\beta$, where $\alpha$ ranges within [1, 2) and $\beta \in \mathbb{Z}$ is an integer. Practically, we limit $\beta$ to the range of $-127 \leq \beta \leq 127$, mirroring the constraints of single precision floating-point arithmetic. Both the sign and $\beta$, as discrete entities, are embedded in $d$ dimensions. The fractional component is segmented into $K$ uniformly distributed bins using a method of soft binning. These embeddings are then also summed up together with the column name embedding.

This numerical encoding enjoys multiple advantageous properties:

\emph{i) Adaptability to Varying Scales:} This method is effective across diverse data scales. For datasets spanning multiple orders of magnitude, the most critical components are the exponent and sign. For more uniformly scaled data, the fractional part gains prominence. Unlike typical normalization methods, both orders of magnitude difference and small scale ones can be captured.

\emph{ii) Outlier Robustness:} Unlike other scaling techniques such as min/max scaling, having one (or many) outliers within a dataset does not compromise the encoding of other rows.

\emph{iii) Versatility for Individual Observations:} The encoding is practical for single data rows, making it suitable both for pre-training on rows provided as sets of key-value pairs and for zero-shot inference. This improves upon methods like quantile encoding, which do not make sense outside a reference dataset and typically require multiple unique values to be present.

\subsection{Backbone}
\ours{} employs a modified version of the standard transformer encoder \cite{vaswani2017attention} as the structural backbone. This model, originally designed for natural language processing tasks, has been adapted to align with the nuances of tabular data. Specifically, with respect the BERT \cite{devlin-etal-2019-bert} implementation, we employ two important modifications. \emph{First}, we have substituted the traditional positional encodings in BERT with column name embeddings. This adaptation allows our model to recognize and interpret the structural attributes of tabular data, where the importance and arrangement of data are dictated by the name of the corresponding columns rather than their sequential order. \emph{Second}, we have omitted the initial \texttt{[CLS]} token that is standard in BERT's architecture. In BERT's typical applications, this token is needed during pre-training for the next sentence classification task. However, as our model does not engage in such tasks, the \texttt{[CLS]} token would never have any loss associated to it, rendering it superfluous.

\subsection{Decoding}\label{sec:decoding}
Decoding transforms the tokens that have gone through the backbone encoder into features similar to those in the encoding. For dates, we only predict day, month, and year. For numbers, we make a small but crucial change: in the fraction part, we do not predict the raw value $\alpha \in [1, 2)$ but predict either $\alpha-1 \in [0, 1)$ or $2-\alpha \in (0, 1]$, depending on the parity of $\beta$. This can be expressed as $\tilde \alpha = (-1)^\beta \big(\alpha-\frac{3}{2}\big) + \frac{1}{2}$. This ensures the target $\tilde \alpha$ is a continuous function of $x$, which has theoretical advantages for computing the derivative of the loss and has been empirically proven to perform better for regression tasks (see Tab.~\ref{tab:ablation_output_number} for details).

\subsection{Training}
Our methodology relies on a multi-task learning objective, incorporating contextually selected loss components based on input type. Specifically, we integrate four distinct loss measures: binary cross-entropy, cross-entropy, Huber loss, and $L^2$ loss, to enhance the model's adaptability and accuracy across different tasks. We employ the following loss configuration:

\textbf{i) Cross-entropy loss} is used for discrete output fields, including day, month, year, sign, and exponent.\\
\textbf{ii) Binary cross-entropy} loss is designated for the fraction part of numerical values, following its piecewise linear transformation detailed in Section \ref{sec:decoding}.\\
\textbf{iii) Huber loss} is applied to predictions of text embeddings.\\
\textbf{iv) $L^2$ loss} is employed for tasks involving a single 1-dimensional head, specifically for regression during fine-tuning.

\noindent{\textbf{Pre-training:}} Pre-training involves masked prediction modeling. Mimicking the original BERT \cite{devlin-etal-2019-bert} and follow-up works (e.g. RoBERTa \cite{RoBERTa}), given a row for pre-training, we select each cell with a fixed probability for masking. Following recent trends (e.g., \cite{wettig2022should}), we use a higher 30\% probability of masking. For each cell selected for masking, we either zero out its content vector (with 80\% probability), leave it unchanged (with 10\% probability), or replace it with a random value (with 10\% probability). Unlike text modeling, in this last case, we restrict the choice to other values sampled from the same column within the same table. This makes it more challenging for the model and solves the issue of selecting a potentially arbitrary, artificial random text or number.

Training the model entails optimizing the following objective:
\begin{equation}
    \theta^* = \arg\min_\theta \sum_{i \in \{d, m, y, s, f, e, t\}} \omega_i\mathcal{L}_{i}(\theta),
\end{equation}
where $\theta\in\mathbb{R}^k$ denotes the $k$ trainable parameters, $\omega_i\in\mathbb{R}$ are the weighting terms, $\mathcal L_i$, for $i \in \{d, m, y, s, f, e\}$ denote the cross-entropy losses for \textbf{d}ay, \textbf{m}onth, \textbf{y}ear, \textbf{s}ign, \textbf{f}raction, and \textbf{e}xponent, and the Huber loss for \textbf{t}ext embeddings, respectively. Each of the 7 losses is the result of averaging across all masked tokens of the corresponding type (if any).

\noindent{\textbf{Fine-tuning:}} During fine-tuning, the decoding heads are removed and replaced by a single ``pooling'' head (e.g., selecting the first token from the transformer output), followed by two linear layers with a non-linearity and dropout in between. For classification, the final layer has as many output nodes as there are classes, trained with cross-entropy loss. For regression, two strategies can be employed. On the one hand, we employ the same shape as the number decoding head above (typically 257 nodes, of which one is used to predict the sign bit, one for $\tilde \alpha \in [0, 1]$, and the rest for the exponent in the $[-127, 127] \cap \mathbb{Z}$ range). In this case, the same losses are applied as for pre-training. Alternatively, a single value can be predicted, and the $L^2$ loss is used. In this last case, it is necessary to normalize the target variable to zero mean and unit variance beforehand.

\section{Experiments \& Results} \label{sec:experiments}
\subsection{Setup} 
For the text embeddings, we leverage the SentenceTransformers~\cite{reimers-2019-sentence-bert} framework and a small open-source LLM for efficiency reasons (\texttt{all-MiniLM-L6-v2})\footnote{\url{https://huggingface.co/sentence-transformers/all-MiniLM-L6-v2}}, which was trained on 1 billion sentences and produces embeddings $v \in \mathbb{R}^{384}$. 

Additionally, we adopt the BERT nomenclature for model sizes (mini, small, medium, base, large), spanning from $N=4$ layers and $d=256$ for ``mini'' to $N=24$ and $d=1024$ for ``large''. Unless otherwise specified, results are reported for the ``base'' configuration of $N=12$ layers and $d=768$. Regarding the weights $\omega_i$, we employ a cascading uniform approach, i.e. $\omega_t = \frac{1}{3}$ and $\omega_i = \frac{1}{9}$ for all other $i \in \{d, m, y, s, f, e\}$.

Our model is pre-trained using tabular data from English Wikipedia, similar to \cite{tapas}. This data consists of \emph{infoboxes} (treated as single-row tables) and \emph{wikidata} (multi-row tables found in article texts)—refer to Tab~\ref{tab:pre-training_data} for a breakdown of data statistics.

\begin{table*}[h]
\centering
\caption{\textbf{Data distribution:} Wikipedia pre-training data.}
\label{tab:pre-training_data}
\begin{tabular}{c|ccccc}
\toprule
Name & \# tables & Avg. rows & Avg. columns & Avg. cells & Total cells\\
\midrule
Infoboxes & 4.2M & 1 & 8.5 & 8.5 & 35M\\
Wikidata & 4.4M & 11.8 & 5.6 & 73 & 320M\\
\bottomrule
\end{tabular}
\end{table*}

\subsection{Training}
\begin{figure}[h]
    \centering
    \includegraphics[width=0.99\textwidth]{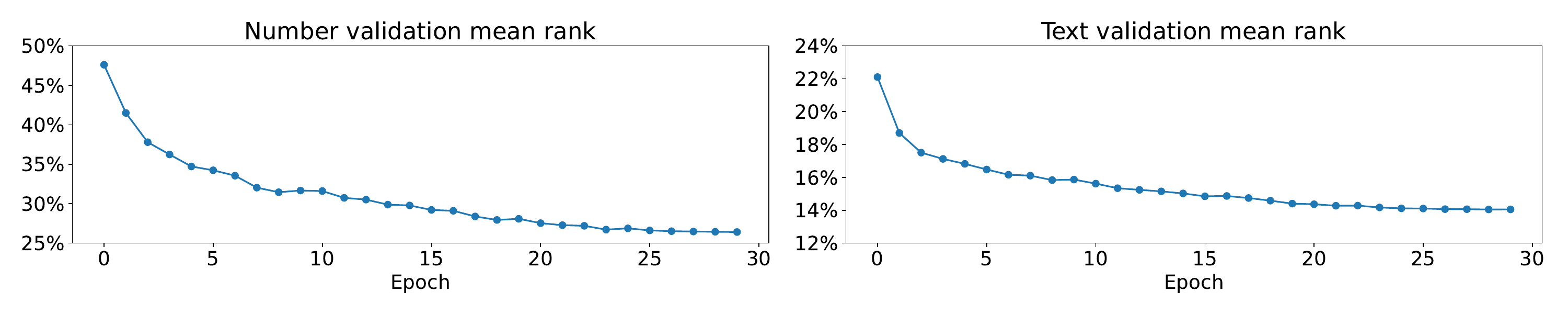}
    \caption{\textbf{Epoch-wise pre-training performance: }Validation metrics per epochs for the number and text head during pre-training.}
    \label{fig:valid_metrics}
\end{figure}

\noindent{\textbf{Pre-training:}} During pre-training, we randomly sample a single row from each table in the corpus each epoch, thus exposing the model to 8.6M rows per epoch across 30 epochs. This amounts to 1.8B tokens sampled from the overall 355M data points. Remark that not all \emph{wikidata} rows are used in training, though \emph{infobox} rows are reviewed 30 times. We use a triangular learning rate schedule peaking at $3 \cdot 10^{-4}$, starting with a warmup over $5\%$ of the training data, followed by a linear decay to zero. A batch size of 4096 is used, and training is conducted on a single NVIDIA A10 GPU with gradient accumulation, with micro batch sizes ranging from 64 to 256 based on model size. \\
For validation, 28k \emph{wikidata} tables were randomly selected, with one row each set aside. Validation involves sequentially masking each cell for a prediction task, totaling a validation set size of 155k. Performance is evaluated using cosine similarity for text and absolute difference for numeric and date fields; similarities between predictions and ground truth are compared to similarities between ground truth and unique values within the same column. The final metric is then the average of relative rankings; results are reported in Fig.~\ref{fig:valid_metrics}.

\noindent{\textbf{Fine-tuning:}} For our study, we utilized a compilation of 51 datasets referenced in \cite{CARTE}, consisting of 11 binary classification and 40 regression datasets. Unlike \cite{CARTE}, we consistently split data into $80\%$ training and $20\%$ testing across all datasets, with results detailed in Tab.~\ref{tab:results}. We also evaluated our model on 45 predominantly numerical datasets from \cite{NEURIPS2022_0378c769}, with results presented in Tab.~\ref{tab:results_numeric}. These datasets were cleaned without strictly adhering to \cite{NEURIPS2022_0378c769} protocols (e.g. multi-class classification was maintained) and limited to $50k$ rows through random sampling, then subject to the same $80\%-20\%$ data split.

Fine-tuning was conducted for up to 100 epochs with early stopping (patience of 20 epochs), typically concluding between 25 and 70 epochs. Similar to \cite{CARTE}, we experimented with ensemble learning via bagging, using 10 models. For regression, we used here a single head and $L^2$ loss, predicting the target value, normalized to zero mean and unit variance. Different types of regression targets are analyzed in Section \ref{sec:ablation}.

\subsection{Results}
We benchmarked our models against robust baselines, including tree boosting algorithms (XGBoost~\cite{Chen:2016:XST:2939672.2939785}, CatBoost~\cite{conf/nips/ProkhorenkovaGV18}) and transformer-based architectures (CM2~\cite{ye2024crosstable}, CARTE~\cite{CARTE}). We also compared against feature-engineered versions of boosting algorithms by providing them with text embeddings, reduced to 50 dimensions via PCA. Text columns with fewer than 50 unique values are treated as categorical and retain default encoding methods (one-hot for XGBoost, target encoding for CatBoost). AutoGluon was included as a baseline. 
Model performance was evaluated using accuracy for classification tasks and $R^2$ score for regression, capping the latter at zero. Note that our results differ from those in~\cite{CARTE} due to different data splits and baselines. \cite{CARTE} used smaller datasets (up to 2048 points), different baselines (CatBoost without text embeddings, XGBoost with different text embeddings), and different metrics (AUC and normalized scores).\\
Analyzing datasets with categorical data from \cite{CARTE}, our model combined with bagging shows state-of-the-art performance in regression and comes close for classification. Results for \ours{}, CARTE, AutoGluon, and CatBoost with embeddings are statistically close, while XGBoost slightly trails. Without bagging, our model outperforms CARTE, CM2, as well as boosting methods which were not enriched via text embeddings. \\
For numerical datasets from \cite{NEURIPS2022_0378c769}, both CatBoost and AutoGluon outperform other models. However, \ours{} is competitive, surpassing XGBoost, CARTE, and CM2. This differentiation highlights our model's adaptability.


\begin{table*}[h]
\centering
 \caption{\textbf{Evaluation of regression and classification tasks:} Performance on 51 benchmarks. \textbf{Top:} Standard approaches. \textbf{Bottom:} Ensemble models with embeddings.
 \textbf{Datasets:} Average scores on 51 datasets from \cite{CARTE}. Training on 80\% of the data, testing on 20\%.
 }
   \label{tab:results}
   
\begin{tabular}{lcccc}
\toprule
Method & Acc. (\%) & Cl. Rank & $R^2$ & Reg. Rank\\
\midrule
CARTE w/o bagging~\citep{CARTE} & 75.2 & 7.1 & 67.6 & 6.8\\
CatBoost~\citep{conf/nips/ProkhorenkovaGV18} & 75.4 & 6.2 & 66.7 & 6.7\\
XGBoost~\citep{Chen:2016:XST:2939672.2939785} & 71.8 & 9.5 & 59.0 & 8.7\\
CM2~\citep{ye2024crosstable} & 76.3 & 6.6 & 4.9 & 10.0\\
\textbf{\ours{} w/o bagging} & \textbf{77.0} & \textbf{5.4} & \textbf{71.4} & \textbf{4.2}\\
\hdashline
CARTE 10 models bagging & 78.3 & 3.7 & 72.3 & 3.2\\
CatBoost + Embeddings & \textbf{78.4} & \textbf{2.6} & 72.3 & 3.5\\
XGBoost + Embeddings & 76.5 & 6.5 & 67.5 & 6.8\\
AutoGluon~\citep{agtabular} & \textbf{78.4} & 3.6 & 72.6 & 3.1\\
\textbf{\ours{} 10 models bagging} & 77.8 & 3.7 & \textbf{73.8} & \textbf{1.9}\\
\bottomrule
\end{tabular}
\end{table*}


\begin{table*}[ht]
\centering
 \caption{\textbf{Evaluation of regression and classification tasks:} Performance on 45 benchmarks. \textbf{Top:} Standard approaches. \textbf{Bottom:} Ensemble models with embeddings.
 \textbf{Datasets:} Average scores on 45 numerical datasets \cite{NEURIPS2022_0378c769}. Training on 80\% of the data, testing on 20\%.
 }
   \label{tab:results_numeric}

\begin{tabular}{lcccc}
\toprule
Method & Acc. (\%) & Cl. Rank & $R^2$ & Reg. Rank\\
\midrule
CARTE w/o bagging~\citep{CARTE} & 71.9 & 6.5 & 67.1 & 5.7\\
CatBoost~\citep{conf/nips/ProkhorenkovaGV18} & \textbf{87.8} & \textbf{2.2} & \textbf{75.5} & \textbf{3.2}\\
XGBoost~\citep{Chen:2016:XST:2939672.2939785} & 84.6 & 4.7 & 72.1 & 4.8\\
CM2~\citep{ye2024crosstable} & 82.0 & 6.4 & 3.2 & 8.0\\
\textbf{\ours{} w/o bagging} & 84.2 & 5.1 & 73.4 & 5.0\\
\hdashline
CARTE 10 models bagging & 72.6 & 4.9 & 68.3 & 4.3\\
AutoGluon~\citep{agtabular} & \textbf{87.5} & \textbf{1.7} & \textbf{77.2} & \textbf{1.8}\\
\textbf{\ours{} 10 models bagging} & 84.5 & 4.5 & 75.3 & 3.1\\
\bottomrule
\end{tabular}

\end{table*}


\subsection{Ablation Study}\label{sec:ablation}

\noindent{\textbf{Target number encoding:}}
In Tab.~\ref{tab:ablation_output_number}, we present the outcomes of various experiments investigating different encoding strategies for the target variable, denoted as $y = \pm \alpha \cdot 2^\beta$, with $\tilde \alpha = (-1)^\beta (\alpha-1.5) + 0.5$. Each model configuration predicts either $y$ directly or a triplet consisting of the sign, $\beta$, and either $\alpha$ or $\tilde \alpha$. For direct $y$ prediction, we explore binning into percentiles with cross-entropy loss or using non-binned values with $L^2$ loss. For triplet predictions, both $\alpha$ and $\tilde \alpha$ are tested as binned or non-binned. In non-binned scenarios, $\alpha-1, \tilde \alpha \in [0, 1]$ are learned via binary cross-entropy loss. Furthermore, the results of applying different normalization techniques to $y$ (no normalization, standard scaler, power transform) are investigated.

\begin{table*}[h!]
\centering
\caption{\textbf{Ablation of regression target encoding}. We report separately the average score of $\max(R^2, 0)$ and the number (out of 40 experiments) of ``failures'', i.e., when $R^2 < 0$ (or when power transforms resulted in infinity / NaN), to prevent negative scores from impacting the average too much. Numbers refer to fine-tuning a pre-trained ``base'' size model, without bagging, and with a shorter patience of 10 epochs.}
\label{tab:ablation_output_number}
\begin{tabular}{c|cccccc}
\toprule
Method & Targets & Binned? & Loss & Normalization & $R^2$ score (\%) & \# Failures\\
\midrule
\ours{} $L^2$ & $y$ & No & $L^2$ & Standard & \textbf{70.9} & 0\\
\ours{} $\tilde \alpha$ & $\pm, \tilde \alpha, \beta$ & No & XE & None & 67.3 & 0\\
Raw $L^2$ & $y$ & No & $L^2$ & None & 58.1 & 0\\
Percentile & $y$ & Yes & XE & None & 63.8 & 0\\
Not continuous & $\pm, \alpha-1, \beta$ & No & XE & None & 57.8 & 4\\
Binned $\tilde \alpha$ & $\pm, \tilde \alpha, \beta$ & Yes & XE & None & 64.6 & 1\\
Continuous $L^2$ & $\pm, \tilde \alpha, \beta$ & No & $L^2$ & None & 63.8 & 1\\
Standard $\tilde \alpha$ & $\pm, \tilde \alpha, \beta$ & No & XE & Standard & 64.1 & 0\\
Power $L^2$ & $y$ & No & $L^2$ & Power & 67.2 & 2\\
Power $\tilde \alpha$ & $\pm, \tilde \alpha, \beta$ & No & XE & Power & 58.0 & 4\\
\bottomrule
\end{tabular}
\end{table*}

\begin{figure}[h!]
\centering
\includegraphics[width=1.00\textwidth]{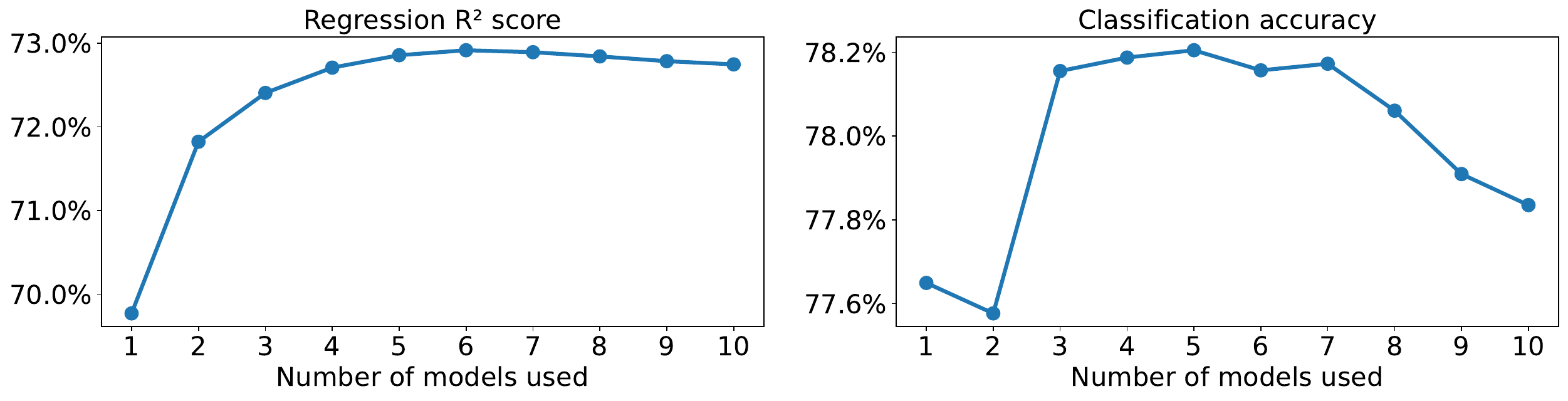}

\includegraphics[width=0.9\textwidth]{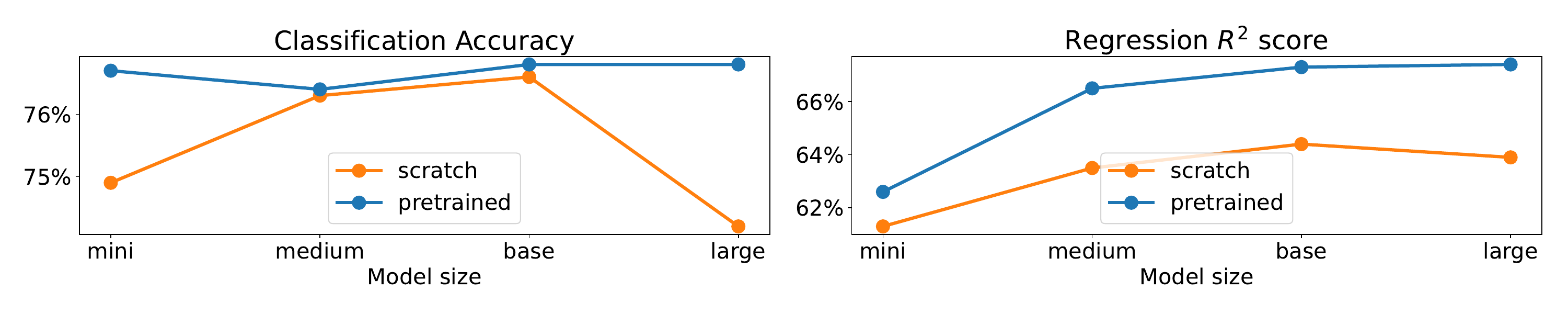}
\caption{\textbf{Performance analysis by model size and count: }\textbf{Top:} Effect of bagging on $R^2$ score and accuracy: this experiment was conducted by selecting top $n$ performing models (based on validation $R^2$ scores and accuracy) from a single batch of 10 runs. \textbf{Bottom:} Performance of models of different sizes, trained on the full training datasets (using patience = 10 epochs and, for regression, predicting $\tilde \alpha$ with binary cross-entropy loss)}
\label{fig:bagging_and_size}
\end{figure}

\noindent{\textbf{Bagging:}} 
In Fig.~\ref{fig:bagging_and_size} (top), we examine the impact of varying the number of top-performing models (in terms of validation metrics) included in bagging on the overall score. Our analysis indicates that averaging results between 5 and 7 models yields the optimal results, increasing overall classification accuracy by $\sim$0.4\% and regression $R^2$ score by $\sim$0.2\%.

Note, although adopting this technique would have significantly improved our results, we refrained from including them in Tab.~\ref{tab:results}, as these ablation results were obtained on the test set directly. However, from an application point of view, it is interesting to note that this approach could maintain, and even surpass, the performance of a full bagging model, while halving the inference runtime.

\noindent{\textbf{Model size and pretraining:}} Finally, we test different model sizes, as well as the effect of pretraining. Results are reported in Figure \ref{fig:bagging_and_size} (bottom). While, on average, pretraining brings a benefit, and larger models do perform better, the relation is noisy. We suspect both effects might become more pronounced with a larger pretraining dataset.

\section{Conclusion and Limitations}\label{sec:conclusions}
This study introduced a novel transformer encoder-based model, \ours{}, which excels on text-heavy datasets and competes strongly with, and occasionally surpasses, established models like XGBoost, CatBoost, and AutoGluon. The model features an innovative encoding structure and a pre-training scheme using masked cell modeling, removing the need for data normalization and enhancing adaptability during fine-tuning, showing potential benefits over traditional tree-based methods.

However, \ours{} faces challenges, possibly stemming from having used Wikipedia data for pre-training, which may not adequately represent diverse tabular datasets. Incorporating broader datasets such as OpenTabs~\cite{ye2024crosstable} or the CSV subset from The Stack~\cite{kocetkov2023the} could increase the model's versatility. Additionally, enhancing the text embedding strategy and optimizing the model for numerical data handling could further improve performance. In particular, larger embedding models have shown better performance in various domains (e.g., \cite{grinsztajn2023vectorizing}), which is a route we have not yet examined in detail.

Moreover, the computational efficiency of \ours{} lags behind more optimized models. Future work should focus on optimizing pre-training strategies to boost efficiency, perhaps by refining larger models or using optimized ensemble configurations. Addressing these limitations is key to improving \ours{}'s effectiveness and real-world applicability, offering promising directions for future research.

\bibliography{refs}
\bibliographystyle{acl_natbib}

\end{document}